\documentclass[10pt,twocolumn,letterpaper]{article}

\usepackage{iccv}
\usepackage{times}
\usepackage{epsfig}
\usepackage{graphicx}
\usepackage{amsmath}
\usepackage{amssymb}

\usepackage{booktabs}
\usepackage{enumitem}
\usepackage{physics}
\usepackage{multirow}
\usepackage{hhline}
\usepackage[page]{appendix}
\DeclareFixedFont{\ttb}{T1}{txtt}{bx}{n}{12} 
\DeclareFixedFont{\ttm}{T1}{txtt}{m}{n}{12}  
\newcommand{\minus}{\scalebox{0.6}[0.8]{$-$}}
\newcommand{\plus}{\scalebox{0.55}[0.75]{$+$}}

\newcommand{\pmm}[1]{p_{m}}
\newcommand{\vsp}{-0.2cm}
\newcommand{\vspt}{-0.2cm}
\newcommand{\vsptt}{-0.25cm}
\usepackage{subcaption}
\usepackage{makecell}
\usepackage{float}
\restylefloat{table}

\usepackage[pagebackref=true,breaklinks=true,colorlinks,bookmarks=false]{hyperref}

\usepackage[breaklinks=true,bookmarks=false]{hyperref}

\iccvfinalcopy 

\def\iccvPaperID{7868} 

\ificcvfinal\fi

\begin{document}

\title{Asymmetric Loss For Multi-Label Classification}

\author{ Emanuel Ben-Baruch\thanks{Equal contribution} \hspace{0.1cm} Tal Ridnik\textsuperscript{*} \hspace{0.1cm} \\ Nadav Zamir \hspace{0.1cm}  Asaf Noy \hspace{0.1cm} Itamar Friedman \hspace{0.1cm} Matan Protter \hspace{0.1cm} Lihi Zelnik-Manor
\vspace{0.5cm} \\ 
DAMO Academy, Alibaba Group\\
{\tt\small $\{$emanuel.benbaruch, tal.ridnik, nadav.zamir, asaf.noy,  itamar.friedman, matan.protter, lihi.zelnik$\}$}\\  {\tt\small @alibaba-inc.com }
}

\maketitle
\ificcvfinal\thispagestyle{empty}\fi

\begin{abstract}
\label{sec:abstract}
In a typical multi-label setting, a picture contains on average few positive labels, and many negative ones. This positive-negative imbalance dominates the optimization process, and can lead to under-emphasizing gradients from positive labels during training, resulting in poor accuracy.
In this paper, we introduce a novel asymmetric loss ("ASL"), which operates differently on positive and negative samples. The loss enables to dynamically down-weights and hard-thresholds easy negative samples, while also discarding possibly mislabeled samples.
We demonstrate how ASL can balance the probabilities of different samples, and how this balancing is translated to better mAP scores.
With ASL, we reach state-of-the-art results on multiple popular multi-label datasets: MS-COCO, Pascal-VOC, NUS-WIDE and Open Images. We also demonstrate ASL applicability for other tasks, such as single-label classification and object detection.
ASL is effective, easy to implement, and does not increase the training time or complexity. \\
Implementation is available at: https://github.com/Alibaba-MIIL/ASL.
\end{abstract}

\section{Introduction}
\label{introduction} 

\begin{figure}\label{fig:intro_figure}
\begin{subfigure}[a]{.2\textwidth}
  \centering
  \includegraphics[width=.9\linewidth]{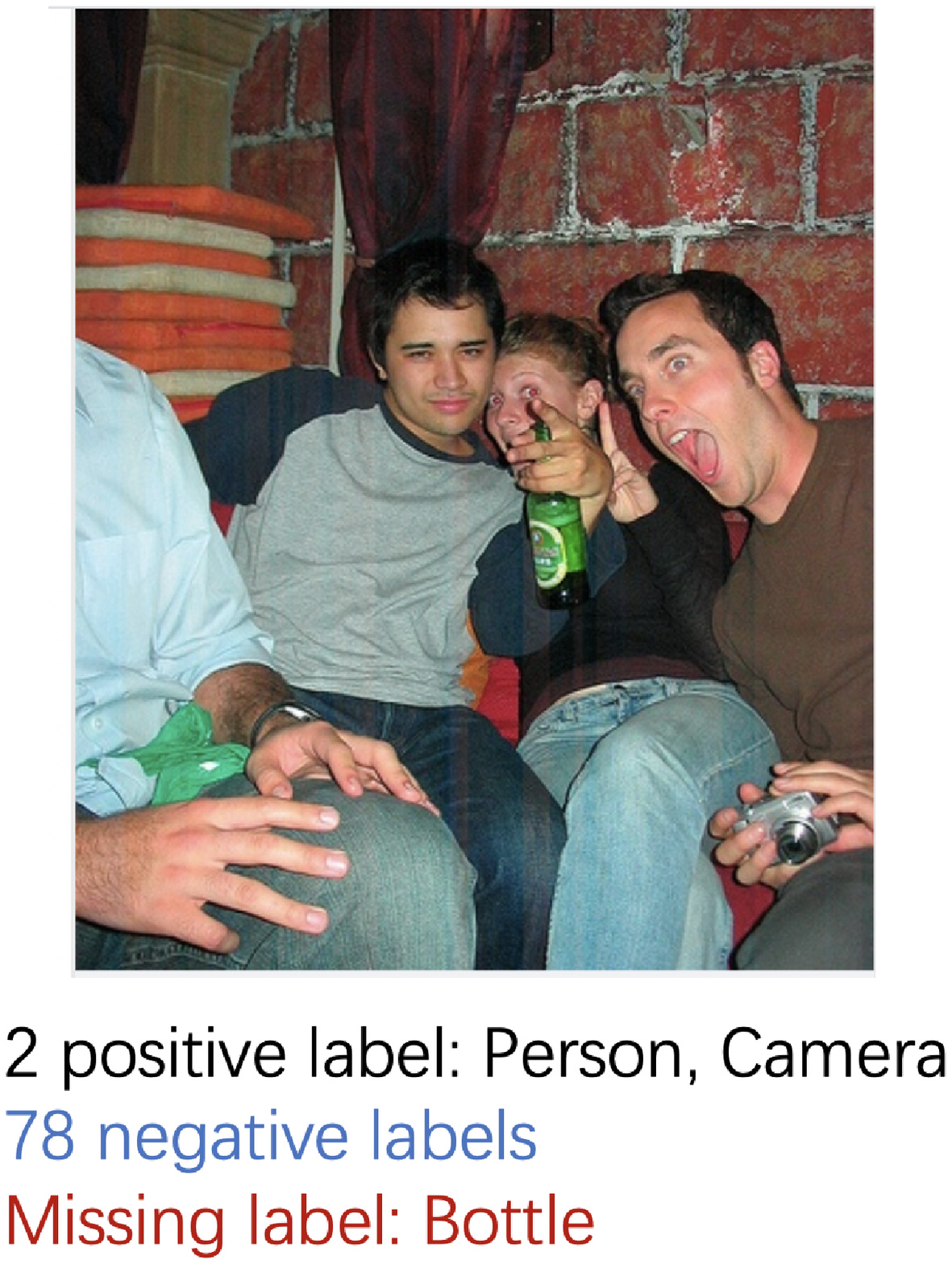}
  \caption{}
  \label{fig:sfig1}
\end{subfigure}%
\begin{subfigure}[h]{.21\textwidth }
  \centering
  \includegraphics[width=1.3\linewidth]{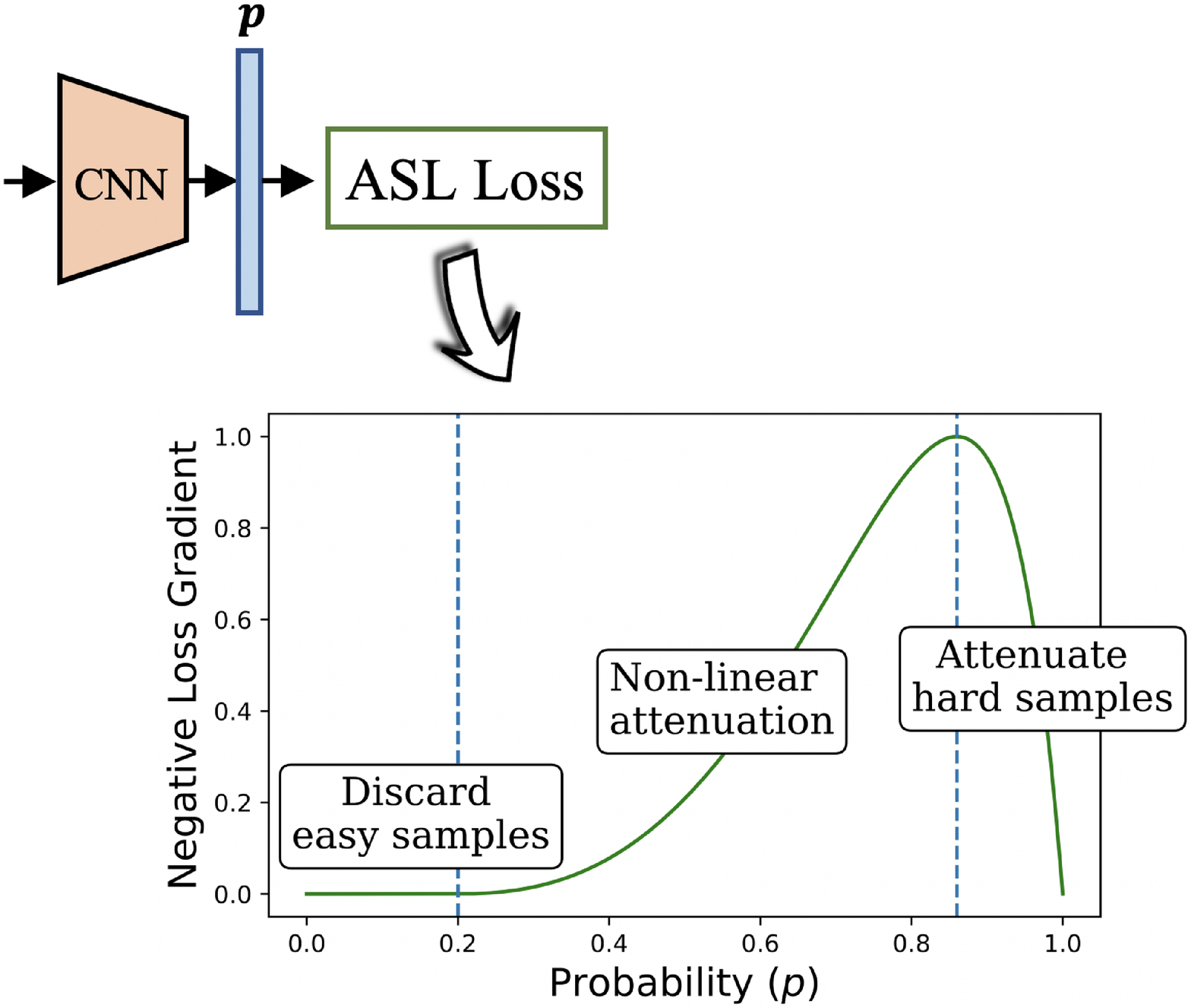}
  \caption{}
  \label{fig:sfig2}
\end{subfigure}
\caption{(a) \textbf{Real world challenges in multi-label classification}. A typical image contains few positive samples, and many negative ones, leading to high negative-positive imbalance. Also, missing labels in ground-truth are common in multi-label datasets. (b) \textbf{Proposed solution with ASL}. 
The loss properties will be detailed in Section \ref{sec:ASL_gradients}}

\label{fig:fig}
\vspace{-3mm}
\end{figure}

Typical natural images contain multiple objects and concepts~\cite{yang2016exploit, yun2021relabel}, highlighting the importance of multi-label classification for real-world tasks. Recently, remarkable advances have been made in multi-label benchmarks such as MS-COCO~\cite{lin2014microsoft}, NUS-WIDE~\cite{chua2009nus}, Pascal-VOC~\cite{everingham2007pascal} and Open Images~\cite{kuznetsova2018open}. 
Notable success was reported by exploiting label correlation via graph neural networks which represent the label relationships~\cite{chen2019multi_MLGCN, chen2019multi, durand2019learning} or word embeddings based on knowledge priors~\cite{chen2019multi_MLGCN, Wang2019MultiLabelCW}. Other approaches are based on modeling image parts and attentional regions~\cite{you2020cross,gao2020multi,wang2017multi,yeattention}, as well as using recurrent neural networks~\cite{nam2017maximizing, wang2016cnn}.

Despite their effectiveness, recent approaches are characterized by extensive architecture modifications and relying on additional external information, such as word embeddings and NLP models. In this work, we question whether such intricate solutions are truly necessary for achieving high performance in multi-label classification tasks.
In particular, we demonstrate that a careful design of the loss function can greatly benefit classification accuracy, while still maintaining a simple and efficient solution, based on standard architectures and training schemes.

A key characteristic of multi-label classification is the inherent positive-negative imbalance created when the overall number of labels is large. Most images contain only a small fraction of the possible labels, implying that the number of positive samples per category will be, on average, much lower than the number of negative samples. 
To address this, \cite{tong2020distribution} suggested a loss function for statically handling the imbalance in multi-label problems. However, it was aimed specifically at long-tail distribution scenarios. 
%
High imbalance is also encountered in dense object detection, where it stems from the ratio of foreground vs. background regions. Some solutions based on resampling methods were proposed, by selecting only a subset of the possible background examples \cite{oksuz2020imbalance}. However, resampling methods are not suitable for handling multi-label classification imbalancing, since each image contains many labels, and resampling cannot change the distribution of only a specific label.

Another common solution in object detection is to adopt the focal loss~\cite{tsung2017focal}, which decays the loss as the label’s confidence increases. This puts focus on hard samples, while down-weighting easy samples, which are mostly related to easy background locations. Surprisingly, focal loss is seldom used for multi-label classification, and cross-entropy is often the default choice (see \cite{chen2019multi_MLGCN,bartlett2008classification,chen2019learning,liu2018multi, gao2020multi}, for example). Since high negative-positive imbalance is also encountered in multi-label classification, focal loss might provide better results, as it encourages focusing on relevant hard-negative samples, which are mostly related to images that do not contain the positive class, but do contain some other confusing categories.
Nevertheless, for the case of multi-label classification, treating the positive and negative samples equally, as proposed by focal loss, is sub-optimal, as it results in the accumulation of more loss gradients from negative samples, and down-weighting of important contributions from the rare positive samples. In other words, the network might focus on learning features from negative samples while under-emphasizing learning features from positive samples.

In this paper, we introduce an asymmetric loss (ASL) for multi-label classification, which explicitly addresses the negative-positive imbalance.
ASL is based on two key properties:
first, to focus on hard negatives while maintaining the contribution of positive samples, we decouple the modulations of the positive and negative samples and assign them different exponential decay factors.
Second, we propose to shift the probabilities of negative samples to completely discard very easy negatives (hard thresholding). By formulating the loss derivatives, we demonstrate that probability shifting also enables to discard very hard negative samples, suspected as mislabeled, which are common in multi-label problems \cite{durand2019learning}.

We compare ASL to the common symmetrical loss functions, cross-entropy and focal loss, and show significant mAP improvement using our asymmetrical formulation. By analyzing the model's probabilities, we demonstrate the effectiveness of ASL in balancing between negative and positive samples. We also introduce a method that dynamically adjusts the asymmetry level throughout the training process, by demanding a fixed gap between positive and negative average probabilities, allowing simplification of the hyper-parameter selection process.

The paper’s contributions can be summarized as follow:
\begin{itemize}[leftmargin=0.4cm]
  \setlength{\itemsep}{0.2pt}
  \setlength{\parskip}{0.2pt}
  \setlength{\parsep}{0.2pt}
    \item We design a novel loss function, ASL, which explicitly copes with two main challenges in multi-label classification: high negative-positive imbalance, and ground-truth mislabeling.
  \item We thoroughly study the loss properties via detailed gradient analysis. An adaptive procedure for controlling the asymmetry level of the loss 
    is introduced, to simplify the process of hyper-parameter selection. 
  \item Using ASL, we obtain state-of-the-art results on four popular multi-label benchmarks. For example, we reach $86.6 \%$ mAP on MS-COCO dataset, surpassing the previous top result by $2.8\%$.
  \item Our solution is effective and easy to use. It is based on standard architectures, does not increase training and inference time, and does not need any external information, in contrast to recent approaches. To make ASL accessible, we share our trained models and a fully reproducible training code.
\end{itemize}


\section{Asymmetric Loss}
In this section, we will first review cross-entropy and focal loss. Then we will introduce the components of the proposed asymmetric loss (ASL), designed to address the inherent imbalance nature of multi-label datasets. We will also analyze ASL gradients, provide probability analysis, and present a method to set the loss' asymmetry levels during training dynamically.

\subsection{Binary Cross-Entropy and Focal Loss}

As commonly done in multi-label classification, we reduce the problem to a series of binary classification tasks.
Given $K$ labels, the base network outputs one logit per label, $z_{k}$. Each logit is independently activated by a sigmoid function $\sigma(z_{k})$. Let's denote $y_{k}$ as the ground-truth for class $k$. The total classification loss, $L_{\text{tot}}$, is obtained by aggregating a binary loss from $K$ labels: 
\begin{equation}\label{eq:loss}
        L_{\text{tot}} = \sum_{k=1}^{K}L\left(\sigma(z_{k}), y_{k}\right).
\end{equation}
A general form of a binary loss per label, $L$, is given by:
\begin{equation}
    L = -y L_{\plus} -(1-y)L_{\minus}
\end{equation}
Where $y$ is the ground-truth label (for brevity we omitted the class index k), and $L_{\plus}$ and $L_{\minus}$ are the positive and negative loss parts, respectively.
Following \cite{tsung2017focal}, focal loss is obtained by setting $L_{\plus}$ and $L_{\minus}$ as:
\begin{equation}\label{eq:FL}
    \begin{cases}
      L_{\plus}= \text{$(1-p)^{\gamma}\log(p)$}\\
      \\
      L_{\minus}= \text{$p^{\gamma}\log(1-p)$}\\
    \end{cases}       
\end{equation}
where $p=\sigma(z)$ is the network's output probability and $\gamma$ is the \textit{focusing parameter}.  $\gamma=0$ yields binary cross-entropy.

By setting $\gamma>0$ in Eq. \ref{eq:FL}, the contribution of easy negatives (having low probability, $p\!\ll\!0.5$) can be down-weighted in the loss function, enabling to focus more on harder samples during training.

\subsection{Asymmetric Focusing}

When using focal loss for multi-label training, there is an inner trade-off: setting high $\gamma$, to sufficiently down-weight the contribution from easy negatives, may eliminate the gradients from the rare positive samples. We propose to decouple the focusing levels of the positive and negative samples. Let $\gamma_{\plus}$ and $\gamma_{\minus}$ be the positive and negative
focusing parameters, respectively. We obtain asymmetric focusing by re-defining the loss:
\begin{equation}\label{eq:ASL-1}
    \begin{cases}
      L_{\plus}=\text{$(1-p)^{\gamma_{\plus}}\log(p)$}\\
      \\
      L_{\minus}=\text{$p^{\gamma_{\minus}}\log(1-p)$}\\
    \end{cases}       
\end{equation}
Since we are interested in emphasizing the contribution of positive samples, we usually set $\gamma_{\minus} > \gamma_{\plus}$.
Asymmetric focusing decouples the decay rates of positive and negative samples. Through this, we achieve better control over the contribution of positive and negative samples to the loss function, and help the network learn meaningful features from  positive samples, despite their rarity. 

It should be noted that methods which address class imbalance via static weighting factors were proposed in previous works \cite{huang2016learning,cui2019class}.
However, \cite{tsung2017focal} found that those weighting factors interact with the focusing parameter, making it necessary to select the two together. In practice, \cite{tsung2017focal} even suggested a weighting factor which favors background samples ($\alpha=0.25$).
In section \ref{sec:Experimental_Study} we will show that simple linear weighting is insufficient to tackle the negative-positive imbalance issue in multi-label classification properly.
For those reasons, we chose to avoid adding static weighting factors to our focusing formulation.

\medskip
\subsection{Asymmetric Probability Shifting}

\begin{figure} [hbt!]
  \centering
  \includegraphics[scale=.55]{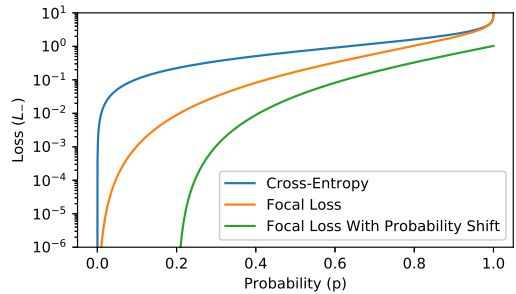}
  \vspace{-0.1cm}
  \caption{\textbf{Loss Comparisons.} Comparing probability-shifted focal loss to regular focal loss and cross-entropy, for negative samples. We used $\gamma_{\minus}=2$ and $m=0.2$.}
  \label{fig:probability_shifted_focal_loss_graph_pdf}
  \vspace{-0.2cm}
\end{figure}

Asymmetric focusing reduces the contribution of negative samples to the loss when their probability is low (soft thresholding). Since the level of imbalancing in multi-label classification can be very high, this attenuation is not always sufficient. Hence, we propose an additional asymmetric mechanism, probability shifting, that performs hard thresholding of very easy negative samples, i.e., it fully discards negative samples when their probability is very low. Let's define the \textit{shifted probability}, $ \pmm] $, as: 
\begin{equation}
\label{eq:p(m)}
\pmm] = \max(p - m, 0)
\end{equation}
Where the \textit{probability margin} $m\!\geq\!0$ is a tunable hyper-parameter. Integrating $p_m$ into $L_{\minus}$ of Eq.\eqref{eq:FL}, we get an asymmetric probability-shifted focal loss: 
\begin{equation} \label{eq:ASL-2}
    L_{\minus}= \text{$(\pmm]) ^{\gamma}\log(1-\pmm] )$}\\
\end{equation}
In Figure \ref{fig:probability_shifted_focal_loss_graph_pdf} we draw the probability-shifted focal loss, for negative samples, and compare it to regular focal loss and cross-entropy.
From a geometrical point-of-view, we can see that probability shifting is equivalent to moving the loss function to the right, by a factor $m$, thus getting  $L_{\minus}\!=\!0$ when $p\!<m\!$. We will later show, via gradient analysis, another important property of the probability shifting mechanism - it can also reject mislabeled negative samples.

Notice that the concept of probability shifting is not limited to cross-entropy or focal loss, and can be used on many loss functions.
Linear hinge loss \cite{bartlett2008classification}, for example, can also be seen as (symmetric) probability shifting of  linear loss. Also notice that logits shifting, as suggested in \cite{tsung2017focal} and \cite{tong2020distribution}, is different from probability shifting due to the non-linear sigmoid operation.

\medskip
\subsection{ASL Definition}
\label{sec:ASL}
To define the Asymmetric Loss (ASL), we integrate the two mechanisms of asymmetric focusing and probability shifting into a unified formula: %
\begin{equation} \label{eq:ASL-3}
  ASL =
    \begin{cases}
      L_{+} = & \text{$(1-p)^{\gamma_{\plus}}\log(p)$}\\
      \\
      L_{-} = & \text{$(\pmm])^{\gamma_{\minus}}\log(1-\pmm])$}\\
    \end{cases}       
\end{equation}
Where $\pmm]$ is defined in Eq.\eqref{eq:p(m)}.
ASL allows us to apply two types of asymmetry for reducing the contribution of easy negative samples to the loss function - soft thresholding via the focusing parameters $\gamma_{\minus}>\gamma_{\plus}$, and hard thresholding via the probability margin $m$.

It can be convenient to set $\gamma_{\plus}=0$, so that positive samples will incur simple cross-entropy loss, and control the level of asymmetric focusing via a single hyper-parameter, $\gamma_{\minus}$. For experimentation and generalizability, we still keep the $\gamma_{\plus}$ degree of freedom.

\subsection{Gradient Analysis}
\label{sec:ASL_gradients}
To better understand the properties and behavior of ASL, we next provide an analysis of the loss gradients, in comparison to the gradients of cross entropy and focal loss. 
Looking at the gradients is useful since, in practice, the network weights are updated according to the gradient of the loss, with respect to the input logit $z$.  The loss gradients for negative samples in ASL are:
\begin{equation} 
\begin{array}{c}
    \begin{split}
      \frac{dL_{\minus}}{dz} & = \frac{\partial L_{\minus}}{\partial p}\frac{\partial p}{\partial z}   \\ 
      & = (\pmm])^{\gamma_{\minus}} \Big[\frac{1}{1-\pmm]}-\frac{   \gamma_{\minus}\log\big(1-\pmm]\big)}{\pmm]}  \Big]p(1-p) 
    \end{split}
\end{array}
\end{equation}
\label{eq:ASL-gradients}
Where $p=\frac{1}{1+e^{-z}}$, and $\pmm]$ is defined in Eq.\eqref{eq:p(m)}.
In Figure \ref{fig:loss_gradients_ASL} we present the normalized gradients of ASL, and compare it to other losses. Following Figure \ref{fig:loss_gradients_ASL}, we can roughly split the negative samples in ASL into three loss-regimes:
\begin{figure} [hbt!]
  \centering
  \includegraphics[scale=.55]{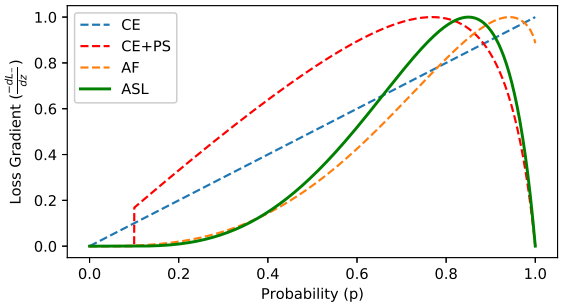}
  \vspace{-0.25cm}
  \caption{ \textbf{Gradient Analysis.} Comparing the loss gradients vs. probability for different loss regimes. CE = Cross-Entropy ($m=\gamma_{\minus}=0$), CE+PS = Cross-Entropy with Probability Shifting ($m>0,\gamma_{\minus}=0$), AF = Asymmetric Focusing ($m=0,\gamma_{\minus}>0$), ASL ($m>0,\gamma_{\minus}>0$).
  }
  \label{fig:loss_gradients_ASL}
  \vspace{\vsptt}
\end{figure} 
\begin{enumerate}[leftmargin=0.4cm]
\setlength{\itemsep}{-3.5pt}
\item 
Hard-threshold - very easy negatives, with $p\!<\!m$, that should be ignored, in order to focus on harder samples.
\item 
Soft-threshold - negative samples, with $p\!>\!m$, that should be attenuated when their probability is low.
\item 
Mislabeled - very hard negative samples, with $p\!>\!p*$, where $p*$ is defined as the point where $\dv{p}(\dv{L}{z})=0$, which are suspected as mislabeled - when the network computes a very large probability for a negative sample, it is possible that the sample was mislabeled, and its correct label should be positive. It has been shown by ~\cite{durand2019learning} that multi-label datasets are prone to mislabeling of negative samples, probably because the manual labeling task is difficult.
When dealing with highly imbalanced datasets, even a small mislabeling rate of negative samples largely impact the training. Hence, rejection of mislabelled samples can be beneficial.
The rejection must be done carefully, to allow the network to propagate gradients from actual misclassified negative examples.
\end{enumerate}
In Table \ref{Table:advanteges} we compare the  properties and abilities of ASL to other losses, according to the gradient analysis. 
\begin{table}[hbt!]
\centering
\setlength{\tabcolsep}{1.9pt}
\small
\begin{tabular}{c|c|c|c|c} 
\hline
                                                                                          & \begin{tabular}[c]{@{}c@{}}Hard\\Threshold\end{tabular} & \begin{tabular}[c]{@{}c@{}}Soft\\Threshold\end{tabular} & \begin{tabular}[c]{@{}c@{}}Discard\\Mislabeled\end{tabular} & \begin{tabular}[c]{@{}c@{}}Continuous\\Gradients \end{tabular}  \\ 
\hline
\hline
CE                                  & -                                                       & -                                                       & -                                                                                & +                                                               \\ 
\hline
AF                              & -                                                       & +                                                       & -                                                                                & +                                                               \\ 
\hline
CE+PS   & +                                                       & -                                                       & +                                                                                & -                                                               \\ 
\hline
ASL (AF+PS) & +                                                       & +                                                       & +                                                                                & +                                                               \\
\hline
\end{tabular}
\caption{\textbf{Properties of different loss} - CE (Cross-Entropy), AF (Asymmetric Focusing), PS (Probability Shifting).}
\label{Table:advanteges}
\vspace{\vsptt}
\end{table}
We can see that only when we combine the two asymmetry mechanisms, focusing and probability margin, we enjoy all the abilities and advantages which are beneficial for imbalanced datasets: hard thresholding of very easy samples, non-linear attenuation of easy samples, rejection of mislabeled samples and continuous loss gradients. 

\subsection{Probability Analysis}
\label{sec:probability_analysis}

In this section, we wish to provide further support to our claim that in multi-label datasets, using a symmetric loss such as cross entropy or focal loss is sub-optimal for learning positive samples' features. We do that by monitoring the average probabilities outputted by the network during the training. This allows us to evaluate the network's level of confidence for positive and negative samples. Low confidence suggests that features were not learned properly. We begin by defining $p_{t}$ as:
\begin{equation} \label{eq:pt}
 p_{t}=\begin{cases} \bar{p} &\text{if $y = 1$}\\ 1 - \bar{p} &\text{otherwise}\end{cases}
\end{equation}  
where $\bar{p}$ denotes the average probability of the samples in a batch at each iteration. Denote by $p_t^{\plus}$ and $p_t^{\minus}$ the average probabilities of the positive and negative samples, respectively, and by $\Delta p$ the probability gap: 
\begin{equation}
    \Delta p=p_t^{\plus}-p_t^{\minus}.
\end{equation}
A balanced training should demonstrate similar level of mean confidence for positive and negative samples, i.e., $\Delta p$ should be small throughout and at the end of the training.
\begin{figure*} [hbt!]
  \centering
  \includegraphics[scale=.54]{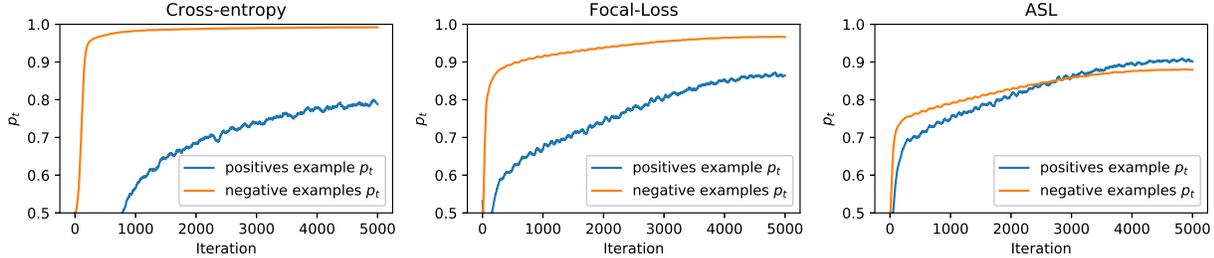}
  \vspace{-0.3cm}
  \caption{\textbf{Probability analysis. } The mean probability of positive and negative samples along the training with cross-entropy, focal loss and ASL, on MS-COCO. For focal loss we used $\gamma=2$. For ASL we used $\gamma_{+}=0$, $\gamma_{-}=2$, $m=0.2$.}
  \label{fig:train_probability_vs_iteration}
  \vspace{-0.25cm}
\end{figure*}

In Figure \ref{fig:train_probability_vs_iteration} we present the average probabilities $p_t^{\plus}$ and $p_t^{\minus}$ along the training, for three different loss functions: cross-entropy, focal loss and ASL. Figure \ref{fig:train_probability_vs_iteration} demonstrates the limitation of using symmetric losses for imbalanced datasets. 
When training with either cross-entropy loss or focal loss, we observe that $p_t^{\minus}\!\gg\!p_t^{\plus}$  (at the end of the training, $\Delta p = -0.23$ and $\Delta p=-0.1$, respectively). This implies that the optimization process gave too much weight to negative samples. 
Conversely, when training with ASL we can eliminate the gap, implying that the network has the ability to properly emphasize positive samples.

Notice that by lowering the decision threshold $p_{th}$ at inference time (a sample will be declared as positive if $p>p_{th}$), we can control the precision vs. recall trade-off, and favor high true-positive rate over low false-negative rate. However, a large negative probability gap, as obtained by the symmetric losses, suggests that the network under-emphasized gradients from positive samples and converged to a local minima, with sub-optimal performances. We will validate this claim in Section \ref{sec:Experimental_Study}.

\subsection{Adaptive Asymmetry}
Hyper-parameters of a loss function are usually  adjusted  via a manual tuning process. This process is often cumbersom,e and requires a level of human expertise. Based on our probability analysis, we wish to offer a simple intuitive way of dynamically adjusting ASL's asymmetry levels, with a single interpretable control parameter.

In the last section we demonstrated that ASL enables to balance a network, and prevent a situation where negative samples have significantly larger $p_t$ than positive samples ($\Delta p<0$).
We now wish to go the other way around, and adjust $\gamma_{\minus}$ dynamically throughout the training, to match a desired probability gap, denoted by $\Delta p_{\text{target}}$. We can achieve this by a simple adaptation of $\gamma_{\minus}$ after each batch, as described in Eq. \ref{dynamics_gamma_minus}.
\begin{equation}
\label{dynamics_gamma_minus}
\gamma_{\minus}\leftarrow \gamma_{\minus}+\lambda (\Delta p-\Delta p_{\text{target}})
\end{equation}
where $\lambda$ is a dedicated step size. As we increase $\Delta p_{\text{target}}$, Eq. \ref{dynamics_gamma_minus} enables us to  dynamically increase the asymmetry level throughout the training, forcing the optimization process to focus more on the positive samples' gradients. Notice that using similar logic to Eq. \ref{dynamics_gamma_minus}, we can also dynamically adjust the probability margin, or simultaneously adjust both asymmetry mechanisms. For simplicity, we chose to explore the case of adjusting only $\gamma_{\minus}$ throughout the training, with $\gamma_{\plus}=0$ and a small fixed probability margin.

Figure \ref{fig:adaptive_gamma} in appendix \ref{sec:adaptive_asymmetry_graph} presents the values of $\gamma_{\minus}$ and $\Delta p$ throughout the training, for $\Delta p_{\text{target}}=0.1$. After $10\%$ of the training, the network converges successfully to the target probability gap, and to a stable value of $\gamma_{\minus}$.
In the next section we will analyze the mAP score and possible use-cases for this dynamic scheme.

\section{Experimental Study}
\label{sec:Experimental_Study}
In this section, we will provide thorough experimentations to better understand the different losses, and demonstrate the improvement we gain from ASL, compared to other losses. We will also test our adaptive asymmetry mechanism, and compare it to a fixed scheme.
For testing, we will use the well-known MS-COCO \cite{lin2014microsoft} dataset (see Section \ref{sec:coco_scores} for full dataset and training details).

\begin{figure} [hbt!]
  \centering
  \includegraphics[scale=.65]{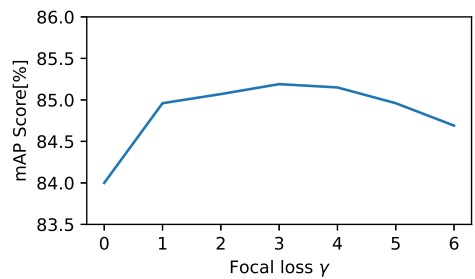}
  \vspace{\vsp}
  \caption{\textbf{mAP Vs. Focal Loss $\gamma$.} Comparing MS-COCO mAP score for different values of focal loss $\gamma$.}
  \label{fig:coco_score_vs_focal_loss_gama}
  \vspace{-0.1cm}
\end{figure}

\textbf{Focal Loss Vs Cross-Entropy:}
In Figure \ref{fig:coco_score_vs_focal_loss_gama} we present the mAP scores obtained for different values of focal loss $\gamma$ ($\gamma=0$ is cross-entropy).
We can see from Figure \ref{fig:coco_score_vs_focal_loss_gama} that with cross-entropy loss, the mAP score is lower than the one obtained with focal loss ($84.0\%$ vs $85.1\%$). Top scores with focal loss are obtained for $2 \leq\gamma\leq4$. With $\gamma$ below that range, the loss does not provide enough down-weighting for easy negative samples. With $\gamma$ above that range, there is too much down-weighting of the rare positive samples.

\begin{figure} [hbt!]
  \centering
  \includegraphics[scale=.65]{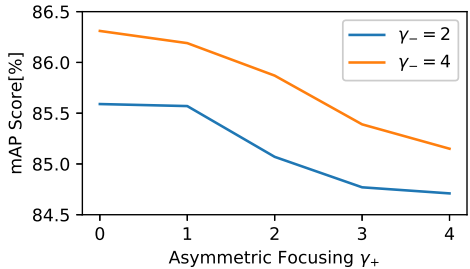}
  \vspace{\vsp}
  \caption{\textbf{mAP Vs. Asymmetric Focusing $\gamma_{+}$.} Comparing MS-COCO mAP score for different value of asymmetric focusing $\gamma_{+}$, for $\gamma_{-}=2$ and $\gamma_{-}=4$.}
  \label{fig:coco_score_vs_gamma_plus}
  \vspace{-0.1cm}
\end{figure}

\textbf{Asymmetric Focusing:}
In Figure \ref{fig:coco_score_vs_gamma_plus} we test the asymmetric focusing mechanism: for two fixed values of $\gamma_{\minus}$, $2$ and $4$, we present the mAP score along the $\gamma_{\plus}$ axis.
Figure \ref{fig:coco_score_vs_gamma_plus} demonstrates the effectiveness of asymmetrical focusing - as we decrease $\gamma_{\plus}$ (hence increasing the level of asymmetry), the mAP score significantly improves. 

Interestingly, simply setting $\gamma_{\plus}=0$ leads to the best results in our experiments. That may further support the importance of keeping the gradient magnitudes high for positive samples. Indeed, allowing $\gamma_{\plus}>0$ may be useful for cases where there is also an abundance of easy positive samples.
Note that we also tried training with $\gamma_{\plus}<0$, to extend the asymmetry further. However, these trials did not converge, therefore they are not presented in Figure \ref{fig:coco_score_vs_gamma_plus}.

\begin{figure} [hbt!]
  \centering
  \includegraphics[scale=.55]{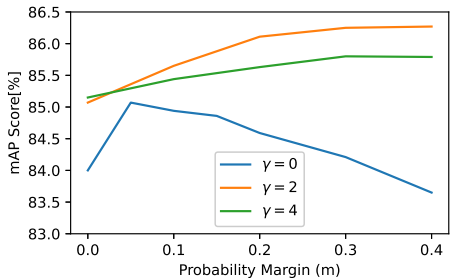}
  \vspace{\vsp}
  \caption{\textbf{mAP Vs. Asymmetric Probability Margin. } Comparing MS-COCO mAP score for different values of asymmetric probability margin, on top of a symmetric focal loss, with $\gamma=0, 2, 4$}
  \label{fig:coco_score_vs_probability_margin}
  \vspace{-0.1cm}
\end{figure}

\textbf{Asymmetric Probability Margin:}
In Figure \ref{fig:coco_score_vs_probability_margin} we apply our second asymmetry mechanism, asymmetric probability margin, on top of cross-entropy loss ($\gamma=0$) and two levels of (symmetric) focal loss, $\gamma=2$ and $\gamma=4$. 

We can see from Figure \ref{fig:coco_score_vs_probability_margin} that both for cross-entropy and focal loss, introducing asymmetric probability margin improves the mAP score. For cross-entropy, the optimal probability margin is low, $m=0.05$, in agreement with our gradient analysis -  cross-entropy with probability margin produces a non-smooth loss gradient, with less attenuation of easy samples. Hence, a small probability margin, which still enables hard threshold for very easy samples and rejection of mislabeled samples, is better. For focal loss, the optimal probability margin is  significantly higher, $0.3\leq  m \leq 0.4$. This again can be explained by analyzing the loss gradients: since focal loss already has non-linear attenuation of easy samples, we need a larger probability margin to introduce meaningful asymmetry. We can also see that when introducing  asymmetric probability margin, better scores are obtained for $\gamma=2$ compared to $\gamma=4$, meaning that asymmetric probability margin works better on top of a modest amount of focal loss.


\textbf{Comparing Different Asymmetries}: Until now we tested each ASL asymmetry separately. In Table \ref{Table:asymmetry_type} we present the mAP scores achieved when combining the asymmetries, and compare them to the scores obtained when applying each asymmetry alone.
Also, we compare ASL results to another asymmetric mechanism - focal loss combined with linear weighting, as proposed in \cite{oksuz2020imbalance}, that statically favors positive samples. The optimal static weight was searched over a range of values between $0.5$ to $0.95$, with skips of $0.05$.
\begin{table}[hbt!]
\centering
\begin{tabular}{|c|c|} 
\hline
Method            &
mAP [\%] \\
\hline
\hline
FL                                                                & 85.1                                                        \\ 
\hline
FL + linear weighting                                             & 85.3                                                       \\ 
\hline
Focusing (ASL)                                                             & 86.3                                                       \\ 
\hline
Probability margin (ASL)                                                     & 86.3                                                       \\ 
\hline
\makecell{Focusing +  Probability margin (ASL)} & \textbf{86.6}     
\\
\hline
\end{tabular}
\caption{\textbf{MS-COCO mAP scores for different asymmetric methods.} Focusing mAP obtained for $\gamma_{+}=0, \gamma_{-}=3$. Margin mAP obtained for $\gamma=2, m=0.3$. Combined mAP obtained for $\gamma_{+}=0, \gamma_{-}=4 , m=0.05.$}
\label{Table:asymmetry_type}

\end{table}

We can see from Table \ref{Table:asymmetry_type} that the best results are obtained when combining the two components of asymmetry. This correlates with our analysis of the loss gradients in Figure \ref{fig:loss_gradients_ASL}, where we demonstrate how combining the two asymmetries enables discarding of very easy samples, nonlinear attenuation of easy samples and rejection of possibly mislabeled very hard negative samples, a result which is not possible when applying only one type of asymmetry.
Table \ref{Table:asymmetry_type} also shows that using static weighing is insufficient to properly handle the high negative-positive imbalance in multi-label classification, and ASL, which operates dynamically on easy and hard samples, performs better.

\textbf{Adaptive Asymmetry:}
We now examine the effectiveness of adjusting the ASL asymmetry levels dynamically, via the procedure proposed in Eq.\ref{dynamics_gamma_minus}. In Table \ref{Table:dynamic_gamma} we present the mAP score, and the final value of $\gamma_{\minus}$, obtained for various values of $\Delta p_{\text{target}}$. 
%
\begin{table}[hbt!]
\centering

\begin{tabular}{|c|c|c|}
\hline
\begin{tabular}[c]{@{}c@{}}$\Delta p_{\text{target}}$\\ \end{tabular} &
$\gamma_{-}$ Final & 
mAP Score [\%] \\ 
\hline
\hline
0                                                                                 & 1.2                                                                               & 85.8                                               \\ 
\hline
0.1                                                                               & 3.3                                                                               & 86.1                                               \\ 
\hline
0.2                                                                               & 5.2                                                                               & \textbf{86.4}                                               \\ 
\hline
0.3                                                                             & 6.2                                                                               & 86.3       
\\
\hline
\end{tabular}
\caption{\textbf{Adaptive Asymmetry}. mAP scores and $\gamma_{-}$ obtained from adaptive asymmetry runs, for different $\Delta p_{\text{target}}$.}
\label{Table:dynamic_gamma}
\end{table}

We can see from Table \ref{Table:dynamic_gamma} that even without any tuning, demanding the unbiased case $\Delta p_{\text{target}}=0$, a significant improvement is achieved compared to focal loss ($85.8\%$ vs. $85.1\%$). Even better scores are obtained when using a higher probability gap, $\Delta p_{\text{target}}=0.2$. Interestingly, extra focus on the rare positive samples ($\Delta p_{\text{target}}>0$) is better than just demanding the unbiased case.

Notice that the top mAP scores obtained from the dynamic scheme are still lower by $0.2\%$ compared to the best ASL score with a fixed $\gamma_{\minus}$.
One possible reason for this (small) degradation is that the training process is highly impacted by the first epochs \cite{aditya2019timematters}. Tuning hyper-parameter dynamically may be sub-optimal at the beginning of the training, which decreases the overall performance. To compensate for the initial recovery iterations, dynamically-tuned $\gamma_{\minus}$ tends to converge to higher values, but the overall score is still somewhat hindered. Due to this decline, we chose to use a fixed asymmetry scheme in section \ref{sec:dataset_results}.

Still, the dynamic scheme can be appealing to a non-expert user, as it allows control of the asymmetry level via one simple interpretable hyper-parameter. In addition, we will explore in the future ways to expand this scheme for other applications, such as tuning $\gamma_{\minus}$ adaptively per class, which can be impractical with a regular exhaustive search.

\section{Dataset Results}
\label{sec:dataset_results}
In this section, we will evaluate ASL on four popular multi-label classification datasets, and compare its results to known state-of-the-art techniques, and to other commonly used loss functions. We will also test ASL's applicability to other computer vision tasks, such as single-label classification and object detection.
\subsection{Multi-Label Datasets}
\subsubsection{MS-COCO}
\label{sec:coco_scores}
MS-COCO \cite{lin2014microsoft} is a widely used dataset to
evaluate computer vision tasks such as object detection, semantic
segmentation and image captioning, and has been adopted recently to evaluate multi-label image classification. For multi-label classification, it contains $122,218$ images
with $80$ different categories, where every image contains on average $2.9$ labels, thus giving an average positive-negative ratio of: $\frac{2.9}{80-2.9}=0.0376$. The dataset is divided to a training set of $82,081$ images
and a validation set of $40,137$ images. Following conventional settings for MS-COCO \cite{Wang2019MultiLabelCW,liu2018multi}, we report the following statistics: mean average precision (mAP), average per-class precision (CP), recall (CR), F1 (CF1) and the average overall precision (OP), recall (OR) and F1 (OF1), for the overall statistics and top-3 highest scores. Among these metrics, mAP, OF1, and CF1 are the main metrics, since they take into account both false-negative and false-positive rates.  

In Table \ref{Table:coco_summary} we compare ASL results to known state-of-the-art methods from the literature, for the main metrics (Full training details and loss hyper-parameters are provided in appendix \ref{sec:general_training_details}). In Table \ref{Table:coco} in appendix \ref{sec:ms_coco_all_common_metrics_appendix} we bring results for all the metrices. 
\begin{table}[hbt!]
\centering
\begin{tabular}{c|c|c|c} 
\hline
Method & mAP            & CF1            & OF1  \\ 
\hline
CADM \cite{chen2019multi}                    & 82.3           & 77.0           & 79.6                       \\
ML-GCN \cite{chen2019multi_MLGCN}                  & 83.0           & 78.0           & 80.3                       \\
KSSNet \cite{liu2018multi}                         & 83.7           & 77.2           & 81.5                       \\
MS-CMA \cite{you2020cross}                  & 83.8           & 78.4           & 81.0                       \\
MCAR  \cite{gao2020multi}                     & 83.8           & 78.0           & 80.3                       \\ 
\hline
ASL (ResNet101)                        & \textbf{85.0}  & \bf{80.3} & \bf{82.3} \\

ASL (TResNet-L)                        & \textbf{86.6}  & \bf{81.4}  & \bf{81.8}              \\
\hline
\end{tabular}
\vspace{\vspt}

\caption{\textbf{Comparison of ASL to state-of-the-art methods on MS-COCO.} All metrics are in \%. Results are reported for input resolution 448.}
\label{Table:coco_summary}

\end{table}

We can see from Table \ref{Table:coco_summary} that using ASL we significantly outperform previous state-of-the-art methods on ResNet101, the commonly used architecture in multi-label classification, and improve the top mAP score by more than $1\%$. Other metrics also show improvement. 

Notice that our ASL-based solution does not require architecture modifications, and does not increase inference and training times. This is in contrast to previous top solutions, which include intricate architecture modifications (attentional regions \cite{gao2020multi}, GCNs \cite{zhang2019bridging}), injecting external data like label embeddings \cite{you2020cross,chen2019multi}, and using teacher models \cite{liu2018multi}. However, ASL is fully complementary to those methods, and employing them as well could lead to further score improvement, at the cost of increasing training complexity and reducing throughput. In addition, we see from Table \ref{Table:coco_summary} that using a newer architecture like TResNet-L, that was designed to match the GPU throughput of ResNet101 \cite{ridnik2020tresnet}, we can further improve the mAP score, while still keeping the same training and inference time. This is another contribution of our proposed solution - identifying that modern fast architectures can give a big boost to multi-label classification, and the common usage of ResNet101 can be sub-optimal.

\begin{figure} [hbt!]
  \centering
  \includegraphics[scale=.65]{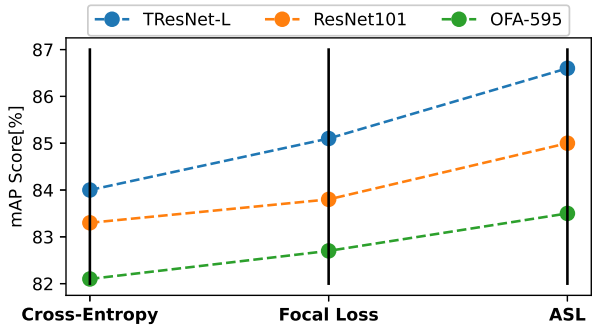}
  \vspace{\vsp}
  \caption{\textbf{Testing different losses on various backbones.}}
  \label{fig:comparing_architectures}
  \vspace{\vsptt}
  \label{Table:loss_different_architecutres}
\vspace{-0.1cm}
\end{figure}

In Figure \ref{Table:loss_different_architecutres} we test the applicability of ASL for different backbones, by comparing the different loss functions on three commonly used architectures: OFA-595 \cite{cai2020once}, ResNet101 and TResNet-L. We can see from Figure \ref{Table:loss_different_architecutres} that on all backbones, ASL outperforms focal loss and cross-entropy, demonstrating its robustness to backbone selection, and its superiority over previous loss functions.

\vspace{-0.1cm}
\vspace{-0.1cm}

\paragraph{Impact of pretraining and input resolutions:} In Table \ref{Table:coco_high_resolution} we compare the mAP results obtained with a standard ImageNet-1K pretraining, and the newer ImageNet-21K pretraining \cite{ridnik2021imagenet21k}. We can see that using better pretraining has dramatic impact on the results, increasing the mAP score by almost $2\%$. We also show in Table \ref{Table:coco_high_resolution} that increasing input resolution from $448$ to $640$ can further improve results.
\begin{table}[hbt!]
\centering
\begin{tabular}{c|c|c|c|c} 
\hline
Method & Architecture & \begin{tabular}[c]{@{}c@{}}Pretrain\\Type\end{tabular} & \begin{tabular}[c]{@{}c@{}}Input\\Resolution\end{tabular} & mAP            \\ 
\hline\hline
ASL    & TResNet-L    & 1K                                                     & 448                                                       & 86.6           \\ 
\hline
ASL    & TResNet-L    & 21K \cite{ridnik2021imagenet21k}                                                    & 448                                                       & 88.4           \\ 
\hline
ASL    & TResNet-L    & 21K \cite{ridnik2021imagenet21k}                                                      & 640                                                       & \textbf{89.8}  \\
\hline
\end{tabular}
\caption{\textbf{Comparison of MS-COCO mAP scores for different ImageNet pretraining schemes, and input resolutions.} All metrics are in \%.}
\label{Table:coco_high_resolution}
\end{table}
\label{results:coco}
\subsubsection{Pascal-VOC}
Pascal Visual Object Classes
Challenge (VOC 2007) \cite{everingham2007pascal} is another popular dataset for
multi-label recognition. It contains images from 20
object categories, with an average of 2.5 categories per image. Pascal-VOC is divided to a trainval set of 5,011 images and a test set of 4,952 images. Our training settings were identical to the ones used for MS-COCO.
Notice that most previous works on Pascal-VOC used simple ImageNet pre-training, but some used additional data, like pre-training on MS-COCO or using NLP models like BERT. For a fair comparison, we present our results once with  ImageNet pre-training, and once with additional pre-train data (MS-COCO) and compare them to the relevant works. Results appear in Table \ref{Table:pascal_voc}.

\begin{table} [hbt!]
\centering
\begin{tabular}{c|c|c} 
\hline
Method        & \begin{tabular}[c]{@{}c@{}}mAP~\\(ImageNet \\ Only Pretrain)\end{tabular} & \begin{tabular}[c]{@{}c@{}}mAP~\\(Extra  \\ Pretrain Data)\end{tabular}  \\ 
\hline
\hline
RNN \cite{wang2017multi} & 91.9                                                                     & -                                                                           \\
FeV+LV \cite{yang2016exploit}       & 92.0                                                                     & -                                                                           \\
SSGRL \cite{chen2019learning}         & 93.4                                                                     & 95.0                                                                        \\
ML-GCN \cite{chen2019multi_MLGCN}        & 94.0                                                                     & -                                                                           \\
BMML \cite{li2020bi}          & -                                                                        & 95.0                                                                        \\ 
\hline
ASL (ResNet101)   & \bf{94.4}   & \bf{95.3}  \\
ASL (TResNet-L)   & \bf{94.6}   & \bf{95.8}  \\
\hline
\end{tabular}
\vspace{\vspt}
\caption{\textbf{Comparison of ASL to known state-of-the-art models on Pascal-VOC dataset.} Metrics are in \%.}
\label{Table:pascal_voc}
\end{table}

%
We can see from Table \ref{Table:pascal_voc} that ASL achieves new state-of-the-art results on Pascal-VOC, with and without additional pre-training. In Table \ref{Table:pascal_voc_appendix} in the appendix we compare different loss functions on Pascal-VOC, showing that ASL outperforms cross-entropy and focal loss.

\subsubsection{NUS-WIDE}
In appendix \ref{sec:nus_wide_appendix} we bring results on another common multi-label dataset, NUS-WIDE \cite{chua2009nus}. Table \ref{Table:nus} shows that ASL again outperforms top previous approaches by a large margin, and reach new state-of-the-art result on NUS-WIDE.


%
\subsubsection{Open Images}
Open Images (v6) \cite{kuznetsova2018open} is a large scale dataset, which consists of $9$ million training images and  $125,436$ test images. It is partially annotated with human labels and machine-generated labels. The scale of Open Images is much larger than previous multi-label datasets such as NUS-WIDE, Pascal-VOC and MS-COCO. Also, it contains a considerable amount of unannotated labels. That allows us to test  ASL on extreme classification \cite{zhang2018deep}, and high mislabeleing scenarios. Full dataset and training details appear in the appendix \ref{sec:open_images_appendix}. To the best of our knowledge, no results for other methods were published yet for v6 variant of Open Images. Hence, we compare ASL only to the other common loss functions in multi-label classification. Yet we hope that our result can serve as a benchmark for future comparisons.

Open Images Results appear in Table \ref{Table:open_images}. We can see from Table \ref{Table:open_images} that ASL significantly outperforms focal loss and cross-entropy on Open Images, demonstrating that ASL is suitable for large datasets and extreme classification cases.
\begin{table}[hbt!]
\centering
\begin{tabular}{c|c|c}
\hline
Method    & micro mAP[\%] & macro mAP[\%]  \\ 
\hline
\hline
CE & 84.8       & 92.0        \\
Focal Loss & 84.9       & 92.2        \\
ASL        & \textbf{86.3}       & \textbf{92.8}       \\
\hline
\end{tabular}
\vspace{\vspt}
\caption{\textbf{Comparison of ASL to focal loss and cross-entropy on Open Images V6 dataset.}}
\label{Table:open_images}
\end{table}
\vspace{-0.15cm}
\subsection{Additional Computer Vision Tasks}
In addition to multi-label classification, we wanted to  test ASL on other relevant computer vision tasks. 
Since fine-grain single-label classification and object detection tasks  usually contain a large portion of background or long-tail cases \cite{bartlett2008classification, herbarium}, and are known to benefit from using focal loss, we chose to test ASL on these  tasks.
In sections \ref{sec:herbarium_appendix} and \ref{sec:object_detection_appendix} in the appendix we show that ASL outperform focal loss on relevant datasets for these additional tasks, demonstrating that ASL is not limited to multi-label classification only.
%

\section{Conclusion}
In this paper, we present an asymmetric loss (ASL) for multi-label classification.
ASL contains two complementary asymmetric mechanisms, which
operate differently on positive and negative samples. By examining ASL derivatives, we gained a deeper understanding of the loss properties. Through network probability analysis, we demonstrate the effectiveness of ASL in balancing between negative and positive samples, and proposed an adaptive scheme that can dynamically adjusts
the asymmetry levels throughout the training. Extensive experimental analysis shows that ASL outperforms common loss functions and previous state-of-the-art methods 
on popular multi-label classification benchmarks, including MS-COCO, Pascal-VOC, NUS-WIDE and Open Images.

{\small
\bibliographystyle{ieee_fullname.bst}
\bibliography{egbib.bib}
}

\clearpage

\appendix
\begin{appendices}

\section{
Adaptive Asymmetry dynamics}
\label{sec:adaptive_asymmetry_graph}
 \begin{figure} [hbt!]
  \centering
  \includegraphics[scale=.7]{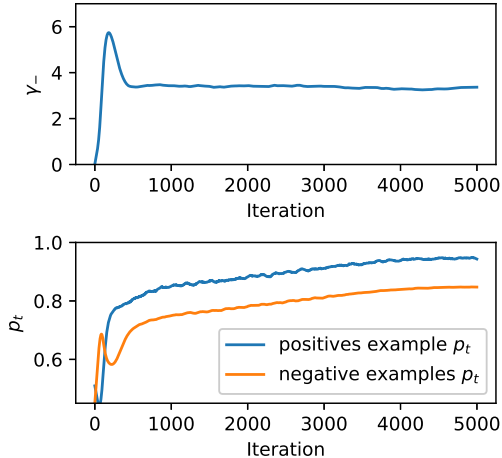}
  \vspace{\vsp}
  \caption{\textbf{Adaptive Asymmetry Dynamics. } Values of $\gamma_{-}$ and $\Delta p$ throughout the training, for $\Delta p_{\text{target}}=0.1$. $\gamma_{+}$ is set to 0, $m$ is set to $0.05$.  }
  \label{fig:adaptive_gamma}
  \vspace{\vsptt}
\end{figure} 

\section{Multi-Label General Training Details}
\label{sec:general_training_details}
Unless stated explicitly otherwise, we used the following training procedure:
We  trained  the  model for $60$ epochs using Adam optimizer  and 1-cycle policy \cite{smith2018disciplined}, with maximal learning rate of $2\text{e-}4$.
For regularization, we used standard augmentation techniques \cite{cubuk2019autoaugment}. We found that the common ImageNet statistics normalization \cite{howard2019searching,cubuk2019autoaugment,tan2019efficientnet} does not improve results, and instead used a simpler normalization - scaling all the RGB channels to be between $0$ and $1$. Following the experiments in section \ref{sec:Experimental_Study}, for ASL we used $\gamma_{\minus}=4$, $\gamma_{\plus}=0$ and $m=0.05$, and for focal loss we used $\gamma=2$.
Our default and recommended backbone for multi-label training is TResNet-L. However, for fair comparison to previous works we also added ResNet101 backbone results on some datasets (TResNet-L and ResNet101 are equivalent in runtime).

\section{Comparing MS-COCO On All Common Metrics}
\label{sec:ms_coco_all_common_metrics_appendix}
In Table \ref{Table:coco} we compare ASL results, to known state-of-the-art methods, on all common metrics for MS-COCO dataset.
\begin{table*}[hbt!]
\setlength{\tabcolsep}{5pt}
\centering
\begin{tabular}{|c||c|c|c|c|c|c|c||c|c|c|c|c|c|} 
\hline
\multicolumn{1}{|c||}{\multirow{2}{*}{Method}} & \multicolumn{7}{c||}{All}                                                                                                       & \multicolumn{6}{c|}{Top 3}                                                                                                                                   \\ 
\cline{2-14}
\multicolumn{1}{|l||}{}                         & mAP            & CP             & CR             & CF1            & OP             & OR             & \multicolumn{1}{l||}{OF1} & CP           & CR           & CF1          & OP           & OR           & OF1           \\ 

\hline
CADM \cite{chen2019multi}                                              & 82.3           & 82.5           & 72.2           & 77.0           & 84.0           & 75.6           & 79.6                      & 87.1                    & 63.6                    & 73.5                     & 89.4                    & 66.0                    & 76.0                      \\
ML-GCN \cite{chen2019multi_MLGCN}                                         & 83.0           & 85.1           & 72.0           & 78.0           & 85.8           & 75.4           & 80.3                      & 87.2                    & 64.6                    & 74.2                     & 89.1                    & 66.7                    & 76.3                      \\
KSSNet \cite{liu2018multi}                                            & 83.7           & 84.6           & 73.2           & 77.2           & 87.8           & 76.2           & 81.5                      & -                       & -                       & -                        & -                       & -                       & -                         \\
MS-CMA \cite{you2020cross}                                          & 83.8           & 82.9           & 74.4           & 78.4           & 84.4           & 77.9           & 81.0                      & 86.7                    & 64.9                    & 74.3                     & 90.9                    & \textbf{67.2}                    & 77.2                      \\
MCAR \cite{gao2020multi}                                                 & 83.8           & 85.0           & 72.1           & 78.0           & 88.0           & 73.9           & 80.3                      & 88.1                    & \textbf{65.5}                    & \textbf{75.1}                     & 91.0                    & 66.3                    & 76.7                      \\ 
\hline
ASL (TResNet-L)                                             & \textbf{86.6} & \textbf{87.2} & \textbf{76.4} & \textbf{81.4} & \textbf{88.2} & \textbf{79.2} & \textbf{81.8}            & \textbf{91.8}            & 63.4            & \textbf{75.1}             & \textbf{92.9}           & 66.4           & \textbf{77.4}              \\
\hline
\end{tabular}
\caption{\textbf{Comparison of ASL to known state-of-the-art models on MS-COCO dataset.} All metrics are in \%. Results are reported for input resolution 448.}
\label{Table:coco}
\end{table*}



\section{Comparing Loss Function on Pascal-VOC Dataset }
\label{sec:pascal_voc_appendix}
In Table \ref{Table:pascal_voc_appendix} we compare ASL results to other loss functions on Pascal-VOC dataset.

\begin{table} [hbt!]
\centering
\begin{tabular}{c|c|c} 
\hline
Method        & \begin{tabular}[c]{@{}c@{}}mAP~\\(ImageNet \\ Only Pretrain)\end{tabular} & \begin{tabular}[c]{@{}c@{}}mAP~\\(Extra  \\ Pretrain Data)\end{tabular}  \\ 
\hline
\hline
CE   & 93.2   & 95.0  \\
Focal Loss   & 93.8   & 95.4  \\
ASL   & \bf{94.6}   & \bf{95.8}  \\
\hline
\end{tabular}
\vspace{\vspt}
\caption{\textbf{Comparison of ASL to other loss functions on Pascal-VOC dataset.} Metrics are in \%.}
\label{Table:pascal_voc_appendix}
\end{table}

\section{NUS-WIDE}
\label{sec:nus_wide_appendix}
NUS-WIDE \cite{chua2009nus} dataset  originally 
contained 269,648 images from Flicker, that have been
manually annotated with 81 visual concepts. Since some urls have
been deleted, we were able to download only 220,000 images, similar to \cite{durand2019learning}.
We can find in previous works \cite{wang2019baseline,liu2018multi} other variants of NUS-WIDE dataset, and its hard to do a one-to-one comparison. We recommend using our publicly available variant\footnote{Our NUS-WIDE variant can be download from: \url{https://drive.google.com/file/d/0B7IzDz-4yH_HMFdiSE44R1lselE/view}} for standardization and a completely fair comparison in future works.
We used the standard 70-30 train-test split \cite{durand2019learning, wang2019baseline,liu2018multi}.  Our training  settings were identical to the ones used for MS-COCO.
\begin{table}[hbt!]
\centering

\begin{tabular}{c|c|c|l} 
\hline
Method & mAP                       & CF1                       & OF1   \\ 
\hline
\hline
S-CLs \cite{liu2018multi}  & 60.1                      & 58.7                      & 73.7  \\
MS-CMA \cite{you2020cross} & 61.4                      & 60.5                      & 73.8  \\
SRN \cite{zhu2017learning}   & 62.0 & 58.5 & 73.4  \\ 
ICME \cite{chen2019multi}   & 62.8 & 60.7 & 74.1  \\ 
\hline
ASL (ResNet101)   & \bf{63.9} & \bf{62.7}  & \bf{74.6}   \\
ASL (TResNet-L)  & \bf{65.2} & \bf{63.6}  & \bf{75.0}   \\
\hline
\end{tabular}
\caption{\textbf{Comparison of ASL to known state-of-the-art models on NUS-WIDE dataset.} All metrics are in \%.}
\label{Table:nus}

\end{table}

%
We can see from Table \ref{Table:nus} that ASL improves the known state-of-the-art results on NUS-WIDE by a large margin.
In Table \ref{Table:nus_appendix} we compare ASL results to other loss functions on NUS-WIDE dataset, again showing that ASL outperform cross-entropy and focal-loss..
\begin{table}[hbt!]
\centering

\begin{tabular}{c|c|c|l} 
\hline
Method & mAP                       & CF1                       & OF1   \\ 
\hline
\hline
CE (Ours)   & 63.1 & 61.7  & 74.6   \\
Focal loss (Ours)   & 64.0 & 62.9  & 74.7   \\
ASL (Ours)   & \bf{65.2} & \bf{63.6}  & \bf{75.0}   \\
\hline
\end{tabular}
\caption{\textbf{Comparison of ASL to known other loss functions on NUS-WIDE dataset.} All metrics are in \%.}
\label{Table:nus_appendix}

\end{table}

\section{Open Images Training Details}
\label{sec:open_images_appendix}

Due to missing links on flicker, we were able to download only $114,648$ test images from Open Images dataset, which contain about $5,400$ unique tagged classes.
For dealing with the partial labeling methodology of Open Images dataset, we set all untagged labels as negative, with reduced weights. Due to the large the number of images, we trained our network for $30$ epochs on input resolution of $224$, and finetuned it for $5$ epochs on input resolution of $448$. Since the level of positive-negative imbalancing is significantly higher than MS-COCO, we increased the level of loss asymmetry: For ASL, we trained with $\gamma_{-}=7, \gamma_{+}=0$. For Focal loss, we trained with $\gamma=4$. Other training details are similar to the ones used for MS-COCO.
\section{Fine-Grain Single-Label Classification Results}

For testing ASL on fine-grain single-label classification, we chose to work on the competitive  Herbarium 2020 FGVC7 Challenge \cite{herbarium}. 
The goal of Herbarium 2020 is to identify vascular plant species from a large, long-tailed collection Herbarium specimens provided by the New York Botanical Garden (NYBG). The dataset contains over 1M images representing over 32,000 plant species. This is a dataset with a long tail; there are a minimum of 3 specimens per species, however, some species are represented by more than a hundred specimens. The metric chosen for the competition is macro F1 score. For Focal loss, we trained with $\gamma=2$. For ASL, we trained with $\gamma_{-}=4, \gamma_{+}=0$.
The metric chosen for the competition is macro F1 score. 
In Table \ref{Table:herbarium} we bring results of ASL on Herbarium dataset, and compare it to regular focal loss.
\begin{table}[hbt!]
\centering
\begin{tabular}{c|c}
\hline
Method    & macro F1 [\%]  \\ 
\hline
\hline
Focal Loss & 76.1           \\
ASL        & \bf{77.6}           \\
\hline
\end{tabular}
\caption{\textbf{Comparison of ASL to focal loss on Herbarium dataset}. Macro-F1 is the competition official metrics. All results are on an unseen private-set. }
\label{Table:herbarium}
\end{table}
%
We can see from Table \ref{Table:herbarium} that ASL outperforms focal loss on this fine-grain single-label classification dataset by a large margin. Note that Herbarium 2020 was a CVPR-Kaggle classification competition. Our ASL test-set score would achieve the 3rd place in the competition, among $153$ teams.

\label{sec:herbarium_appendix}
\section{Object Detection Results}
For testing ASL on object detection, we used the MS-COCO \cite{lin2014microsoft} dataset (object detection task), which contains a training set of 118k images, and an evaluation set of 5k images. 
For training, we used the popular mm-detection \cite{mmdetection} package, with the enhancements discussed in ATSS \cite{zhang2019bridging} and FCOS \cite{tian2019fcos} as the object detection method. We trained a TResNet-M \cite{ridnik2020tresnet} model with SGD optimizer for $70$ epochs, with momentum of $0.9$ , weight decay of $0.0001$ and batch size of 48.  We used learning rate warm up, initial learning rate of 0.01 and 10x reduction at epochs 40, 60.  For ASL we used $\gamma_{+}=1, \gamma_{-}=2$. For focal loss we used the common value, $\gamma=2$ \cite{tsung2017focal}. Note that unlike multi-label and fine-grain single-label classification datasets, for object detection $\gamma_{+}=0$ was not the optimal solution. The reason for this might be  the need to balance the contribution from the $3$ losses used in object detection (classification, bounding box  and centerness). We should further investigate this issue in the future.

Our object detection method, FCOS \cite{tian2019fcos},  uses $3$ different types of losses: classification (focal loss), bounding box (IoU loss) and centerness (plain cross-entropy). The only component which is effected by the large presence of background samples is the classification loss. Hence, for testing we replaced only the classification focal loss with ASL.

In Table \ref{Table:object_detection} we compare the mAP score obtained from ASL training to the score obtained with standard focal loss.
We can see from Table  \ref{Table:object_detection} that ASL outscores regular focal loss, yielding an 0.4\% improvement to the mAP score.
 \begin{table}[h!]
\centering
\begin{tabular}{c|c}
\hline
Method    &  mAP [\%]  \\ 
\hline
\hline
Focal Loss & 44.0       \\
ASL        & \textbf{44.4}       \\
\hline
\end{tabular}
\caption{\textbf{Comparison of ASL to focal loss on MS-COCO detection dataset.} }
\label{Table:object_detection}
\end{table}

\label{sec:object_detection_appendix}
\end{appendices}

\end{document}